\documentclass[times,twocolumn,final]{elsarticle}
\usepackage[english]{babel}
\usepackage{amssymb}
\usepackage{latexsym}
\usepackage{eqnarray,amsmath}
\usepackage{url}
\usepackage{xcolor}
\usepackage{hyperref}
\usepackage{amsmath}
\usepackage{booktabs}
\usepackage{ulem}
\usepackage{multirow}
\usepackage{threeparttable}
\usepackage{color}
\usepackage{natbib}
\usepackage[ruled]{algorithm2e}
\definecolor{newcolor}{rgb}{.8,.349,.1}

\journal{Medical Image Analysis}
\begin{document}
\begin{frontmatter}
\title{SCPM-Net: An Anchor-Free 3D Lung Nodule Detection Network using Sphere Representation and Center Points Matching}
\author[1,2]{Xiangde Luo}
\author[2]{Tao Song\corref{eq}}
\cortext[eq]{Equal Contribution, X. Luo did this work during his internship at SenseTime.}
\author[1]{Guotai Wang\corref{mycorrespondingauthor}}
\cortext[mycorrespondingauthor]{Corresponding author}
\ead{guotai.wang@uestc.edu.cn}
\author[3]{Jieneng Chen}
\author[2,4]{Yinan Chen}
\author[4]{Kang Li}
\author[5]{Dimitris N. Metaxas}
\author[1,2]{Shaoting Zhang}
        
\address[1]{School of Mechanical and Electrical Engineering, University of Electronic Science and Technology of China, Chengdu, China}

\address[2]{SenseTime Research, Shanghai, China}

\address[3]{College of Electronics and Information Engineering, Tongji University, Shanghai, China.}

% \address[4]{Department of Radiology, Tianjin Medical University Cancer Institute and Hospital, Tianjin, China.}

% \address[5]{with Department  of Radiology, the First Affiliated Hospital of Shandong First Medical University, Jinan, China.
% }
\address[4]{West China Biomedical Big Data Center, Sichuan University
West China Hospital, Chengdu, China.}
\address[5]{ Department of Computer Science, Rutgers University, Piscataway, NJ 08854 USA}
\begin{abstract}
Automatic and accurate lung nodule detection from 3D Computed Tomography (CT) scans plays a vital role in efficient lung cancer screening. Despite the state-of-the-art performance obtained by recent anchor-based detectors using Convolutional Neural Networks (CNNs) for this task, they require predetermined anchor parameters such as the size, number, and aspect ratio of anchors, and have limited robustness when dealing with lung nodules with a massive variety of sizes. To overcome these problems, we propose a 3D sphere representation-based center-points matching detection network (SCPM-Net) that is anchor-free and automatically predicts the position, radius, and offset of nodules without manual design of nodule/anchor parameters. The SCPM-Net consists of two novel components: sphere representation and center points matching. First, to match the nodule annotation in clinical practice, we replace the commonly used bounding box  with our proposed bounding sphere to represent nodules  with the centroid, radius, and local offset in 3D space. A compatible sphere-based intersection over-union loss function is introduced to train the lung nodule detection network stably and efficiently. Second,  we empower the network anchor-free by designing a positive center-points selection and matching (CPM) process, which naturally discards pre-determined anchor boxes. An online hard example mining and re-focal loss subsequently enable the CPM process to be more robust, resulting in more accurate point assignment and mitigation of class imbalance. In addition, to better capture spatial information and 3D context for the detection, we propose to fuse multi-level spatial coordinate maps with the feature extractor and combine them with 3D squeeze-and-excitation attention modules. Experimental results on the LUNA16 dataset showed that our proposed SCPM-Net framework achieves superior performance compared with existing anchor-based and anchor-free methods for lung nodule detection with the average sensitivity at 7 predefined FPs/scan of $89.2\%$. Moreover, our sphere representation is verified to achieve higher detection accuracy than the traditional bounding box representation of lung nodules. Code is available at:~\url{https://github.com/HiLab-git/SCPM-Net}. 
\end{abstract}

\end{frontmatter}
% \linenumbers
\section{Introduction}
\par Lung cancer is one of the leading life-threatening cancer around the world, and the diagnosis at an early stage is crucial for the best prognosis~\citep{siegel2015cancer}. As one of the essential computer-aided diagnosis technologies, lung nodule detection from medical images such as Computed Tomography (CT) has been increasingly studied for automatic screening and diagnosis of lung cancer. Detecting pulmonary nodules is very challenging due to the large variation of nodule size, location, and appearance. With the success of Convolutional Neural Networks (CNN) for object detection, CNN-based algorithms have been widely used for detecting pulmonary nodules from CT scans~\citep{ding2017accurate, dou2017automated, zhu2018deeplung, zheng2019automatic,wang2020focalmix,mei2021sanet,tang2019nodulenet}.
\par Current automatic CNN-based 2D object detection approaches are often based on pre-defined anchors and can be mainly summarized into two categories. The first uses a two-stage strategy, where the first stage obtains a set of region proposals and the second stage classifies each proposal and regresses the bounding box of the associated instance of that proposal. For example, Faster RCNN~\citep{ren2015faster} and its variants~\citep{ding2017accurate, dou2017automated} use a network to obtain the region proposals following a feature extractor in the first stage and employ a regression head and a classification head in the second stage. The second category only uses one stage to achieve a trade-off between inference speed and accuracy, like SSD~\citep{liu2016ssd}, YOLO~\citep{redmon2017yolo9000}. Compared with one-stage detectors, the two-stage detectors use a more complex region proposal network and achieve better performance but also require more inference time. Although these anchor-based methods such as Faster RCNN~\citep{ren2015faster}, RetinaNet~\citep{lin2017focal}, SSD~\citep{liu2016ssd} and YOLO~\citep{redmon2017yolo9000} have achieved promising results in object detection from 2D natural images, they are faced with several limitations when applied to 3D medical images. First, they typically need a very large set of anchor boxes, e.g. more than $100k$ in 2D RetinaNet~\citep{lin2017focal}, resulting in huge computational and memory consumption for 3D medical images. Second, the use of anchor boxes introduces many hyper-parameters and design choices, including the number, size, and aspect ratio of the anchor boxes. Manual design of these hyper-parameters is not only time-consuming but also subject to human experience, which may limit the detection performance. It was indicated in~\citep{ding2017accurate} that detection of small objects is sensitive to the manual design of anchors. Moreover, the default anchor configuration is ineffective for detecting lesions with a small size and large aspect ratio~\citep{zlocha2019improving}, let alone the more complex pulmonary nodules, of which the size can vary by as much as ten times. Therefore, a novel detection framework for objects in 3D medical images with higher efficiency and less manual configuring is highly desirable.

\par Recently, anchor-free one-stage detection methods have attracted many researchers' attention and achieved tremendous success in natural image detection~\citep{tian2019fcos, duan2019centernet, zhou2019objects,wang2020centernet3d,chen2020object}. These anchor-free detectors use key points (i.e., center points~\citep{duan2019centernet}, corner points and extreme points~\citep{law2018cornernet,zhou2019objects}) and offset in different directions to represent the object. However, they may have limited performance when dealing with complex cases. For example, the center point may not be inside the target object, and corner/extreme points may be insufficient to represent objects with irregular shapes. These challenges limit the performance of detector to predict the target's location. Meanwhile, there is a lack of existing works on using 3D anchor-free detectors to boost the accuracy and efficiency of object detection from medical volumetric images.

Besides, almost all existing objection detection methods aim to predict or regress the bounding box location and size directly, because all objects are annotated with dense bounding boxes in natural images~\citep{lin2014microsoft, everingham2010pascal}. However, in clinical pulmonary nodule diagnosis, clinicians pay more attention to the location and diameter of a nodule~\citep{macmahon2017guidelines}. Therefore, a 3D lung nodule is often represented by a sphere by specifying its location and radius~\citep{setio2017validation}, rather than a 3D bounding box. Motivated by these observations, we investigate using sphere (parameterized by the central coordinate and radius) to precisely represent lung nodule in 3D space. However, to the best of our knowledge, there have been no studies on using bounding sphere for 3D nodule representation and detection.

Based on these observations, we propose a novel 3D Sphere Representation-based Center-Points Matching Network (SCPM-Net) for pulmonary nodule detection in 3D CT scans. The SCPM-Net is a one-stage anchor-free detection method, which predicts the probability of a pixel $i$ in a coordinate grid being around the center of a nodule, and simultaneously regresses the radius and offset from pixel $i$ to the real nodule center. It is worth mentioning that few works focus on 3D anchor-free detection in medical images, and unlike recent anchor-free methods~\citep{law2018cornernet, duan2019centernet} for 2D natural images, we don't need a key points estimation network to generate a heatmap. Due to the huge variations of size, shape, and location of lung nodules, using a single centroid to represent the lung nodule may lead to a low sensitivity and further limit the potential for clinical application~\citep{song2020cmpnet}. In addition, we proposed a novel center points matching strategy, which uses $K$ points near the centroid to represent a nodule to improve the sensitivity of detectors. Another highlight of our work is that we introduce a sphere representation for lung nodules in 3D space, and further derive an effective sphere-based loss function ($L_{SIoU++}$) to train our anchor-free detector. Compared to the traditional Intersection-over-Union loss~\citep{zhou2019iou} that is based on bounding boxes, the $L_{SIoU++}$ here is closer to clinical scene,  since a nodule's geometry structure  is similar to a sphere or an ellipsoid in most situations. Moreover, our SCPM-Net is a one-stage end-to-end network without additional false positive reduction module adopted in~\citep{ding2017accurate, dou2017automated,mei2021sanet,tang2019nodulenet}, so that it has a higher efficiency and lower memory cost. 

A preliminary version of this work was published in MICCAI2020~\citep{song2020cmpnet}, where we first introduced a center-points-matching-based anchor-free detection network for lung nodule detection in 3D CT scans. In this extend version, we provide more details of our proposed method and describe this work more precisely. Additionally, we introduce a novel sphere representation-based intersection-over-union loss ($L_{SIoU++}$) and provide more experiments to better demonstrate the effectiveness of $L_{SIoU++}$ for lung nodule detection. Meanwhile, we provide a deeper analysis and discussion of this work. Our main contributions are summarized as follows:
\begin{itemize}
    \item[1)] We mitigate the ineffectiveness of current anchor-based detectors~\citep{ren2015faster, lin2017focal} by discarding pre-determined anchor boxes, as we instead predict a center-point map directly via a points matching strategy. To this end, we propose an anchor-free center-points matching network named as CPM-Net with novel attentive modules~\citep{hu2018squeeze}, online hard example mining~\citep{shrivastava2016training}, and re-focal loss~\citep{lin2017focal}.
    \item[2)]We present the first attempt to represent the pulmonary nodule as a bounding sphere in 3D space. Based on sphere representation, we further derive an effective sphere-based intersection-over-union loss function ($L_{SIoU++}$) to train CPM-Net for pulmonary nodule detection, where we called it SCPM-Net.
    \item[3)]We evaluate our approach on LUNA16 dataset~\citep{setio2017validation}, and experimental results showed that the proposed SCPM-Net achieved superior performance compared with anchor-based and existing anchor-free methods for lung nodule detection from 3D CT scans.
\end{itemize}
\begin{figure*}
\setlength{\abovecaptionskip}{0.cm}
    \centering
    \includegraphics[width=0.97\textwidth]{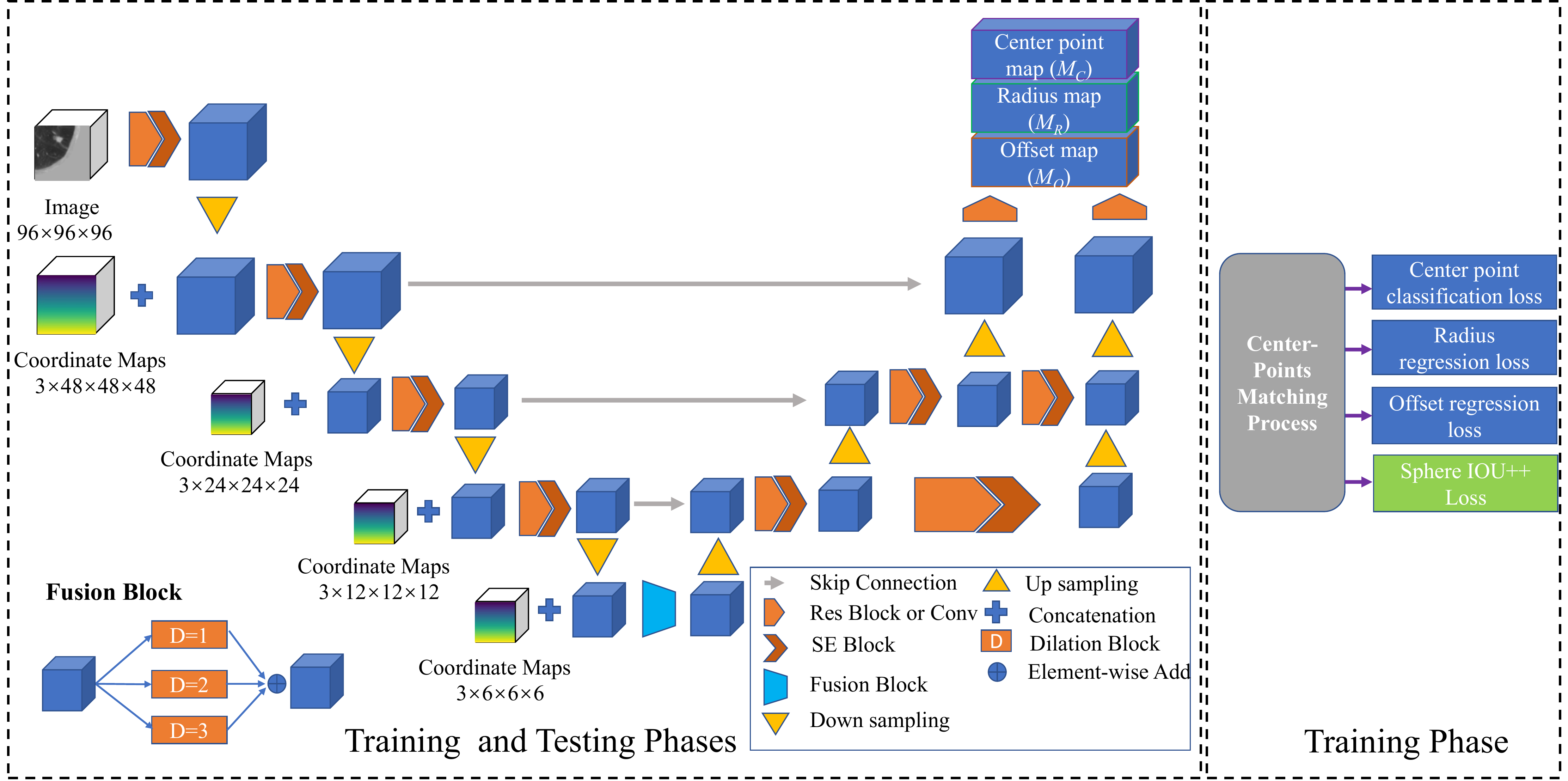}
    \caption{Overview of the proposed SCPM-Net. It consists of an encoder-decoder network with some fusion and attention blocks. The SCPM-Net predicts a center point map $M_C$, a radius map $M_R$ and an offset map $M_O$ to locate target objects without any predefined anchor boxes. In the training stage, a novel center points matching strategy is used to mining hard example for better performance. A geometric-aware sphere representation is proposed to represent the nodule and a sphere-based IoU loss function ($L_{SIoU++}$) is proposed to combine geometric measures with CNN for lung nodule detection.}
    \label{fig:framework}
\end{figure*}
\section{Related works}
\subsection{Anchor-Free Model for Object Detection}Object detection is a fundamental task in the computer vision community and has been studied for many years~\citep{zhao2019object}. In the past few years, many anchor-based frameworks have achieved good performance in natural image detection~\citep{ren2015faster, lin2017focal,redmon2017yolo9000, liu2016ssd}, but they are also limited by high computational cost and long inference time. Differently from these anchor-based detection frameworks, anchor-free methods do not use predefined anchor boxes to predict bounding boxes and have been widely used in natural image detection tasks to boost inference speed and alleviate computational cost~\citep{law2018cornernet, duan2019centernet, zhou2019bottom, tian2019fcos,wang2020centernet3d}. For example, CornerNet~\citep{law2018cornernet} proposed to use a pair of key points to represent the objects and introduced a corner pooling strategy to localize corner points, which reduces the efforts to design anchor boxes. After that, many works~\citep{duan2019centernet, zhou2019bottom, tian2019fcos,wang2020centernet3d} extended the key points-based anchor free detection frameworks in different aspects, e.g., ~\citet{zhou2019bottom} proposed to use extreme points and center points to locate objects. CenterNet~\citep{duan2019centernet} used triplet of key points and the center pooling strategy to capture more recognizable information from the central regions and corner regions.~\citet{zhou2019objects} and~\citet{wang2020centernet3d} detected objects by key-point estimation with regressing the object's size and orientation. Following these methods,~\citet{tian2019fcos} proposed a simple yet efficient anchor-free detection approach called ``FCOS" and proved that a fully convolutional one-stage object detector with centerness can also achieve state-of-the-art performance. In this work, we introduce a novel framework based on sphere representation for anchor-free 3D lung nodule detection. The difference between this work and existing methods are three-fold: First, we represent the lung nodule using the bounding sphere rather than the bounding box in the 3D space, which is a clinically knowledge-driven method~\citep{sladoje2005measurements}. We further extend our proposed sphere representation to a geometric-aware sphere intersection-over-union loss function to impose more geometric constraints over the network in the training stage. Second, we further introduce the squeeze-and-excitation attention and coordinate attention to enhance the network presentation ability. Thirdly, we propose a center points matching strategy to boost the detection sensitivity.

\subsection{Loss function for Object Detection}
Many detection methods used $L_1$ or $L_2$ distance loss functions for bounding box regression~\citep{zhou2019iou}, but they are sensitive to various scales. As a popular evaluation metric, the Intersection-over-Union (IoU) has been used to train various detectors, as it is invariant to the scale~\citep{zheng2020distance,yu2016unitbox}. For two bounding boxes $B^a$ and $B^b$, the IoU loss ($L_{IoU}$) is defined as:
\begin{equation}L_{IoU}=1.0 - \frac{|B^{a} \cap B^{b}|}{|B^{a} \cup B^{b}|}\end{equation}After that, many extensions are proposed to boost the detector's performance by adopting the classical $L_{IoU}$. For example, GIoU~\citep{rezatofighi2019generalized} loss was proposed to deal with gradient vanishing in non-overlapping cases. To further improve models' ability of bounding box regression, DIoU~\citep{zheng2020distance} loss modified $L_{IoU}$ by bringing in the normalized distance between a predicted box and the target box. Another challenge of object detection is the class imbalance problem, a common solution for this problem is using the weighting factor to balance the impact of each class or object. After that, ~\citet{lin2017focal} proposed a  simple yet powerful ``Focal Loss" to address the class imbalance problem by automatically weighting a class based on its hardness. All of these losses have achieved promising performance in 2D nature image detection tasks. Meanwhile, $L_{IoU}$ and ``Focal Loss" are also widely-used in 3D medical image analysis as powerful and efficient loss functions to train 3D detectors~\citep{li2019lung, wang2020focalmix,tang2019nodulenet,mei2021sanet}.

\subsection{3D Pulmonary Nodule Detection}Pulmonary nodule detection plays an essential role in lung cancer diagnosis and treatment. Thus, automated Computer-Aided Diagnosis (CAD) for pulmonary nodule detection has been an active research field. Most of the CAD systems include two stages: nodule candidate detection and false positive reduction. In order to detect nodules at different scales, most previous works extended 2D anchor-based detectors to handle 3D nodule detection, such as using Faster R-CNN~\citep{ren2015faster}, FPN~\citep{LinFeature}. Even so, how to guarantee a high sensitivity with a lower false positive rate is a challenging problem. Hence, False Positive Reduction (FPR) models~\citep{ding2017accurate, dou2017automated} were adopted in pulmonary nodule CAD systems as a second stage to reduce false positives.
To alleviate foreground-background class imbalance, focal loss~\citep{lin2017focal} was used to train anchor-based detectors~\citep{wang2020focalmix}. These methods have largely improved the performance of pulmonary nodule detection~\citep{ding2017accurate,zheng2019automatic,wang2020focalmix}. However, limited by the predefined anchor boxes and multiple stages, these methods need researchers to carefully design many hyper parameters manually to obtain a good performance. In addition, anchor-based multi-stages detectors require higher computational cost and time than anchor-free detectors~\citep{tian2019fcos}. 

\begin{figure*}
\setlength{\abovecaptionskip}{0.cm}
    \centering
    \includegraphics[width=1.0\textwidth]{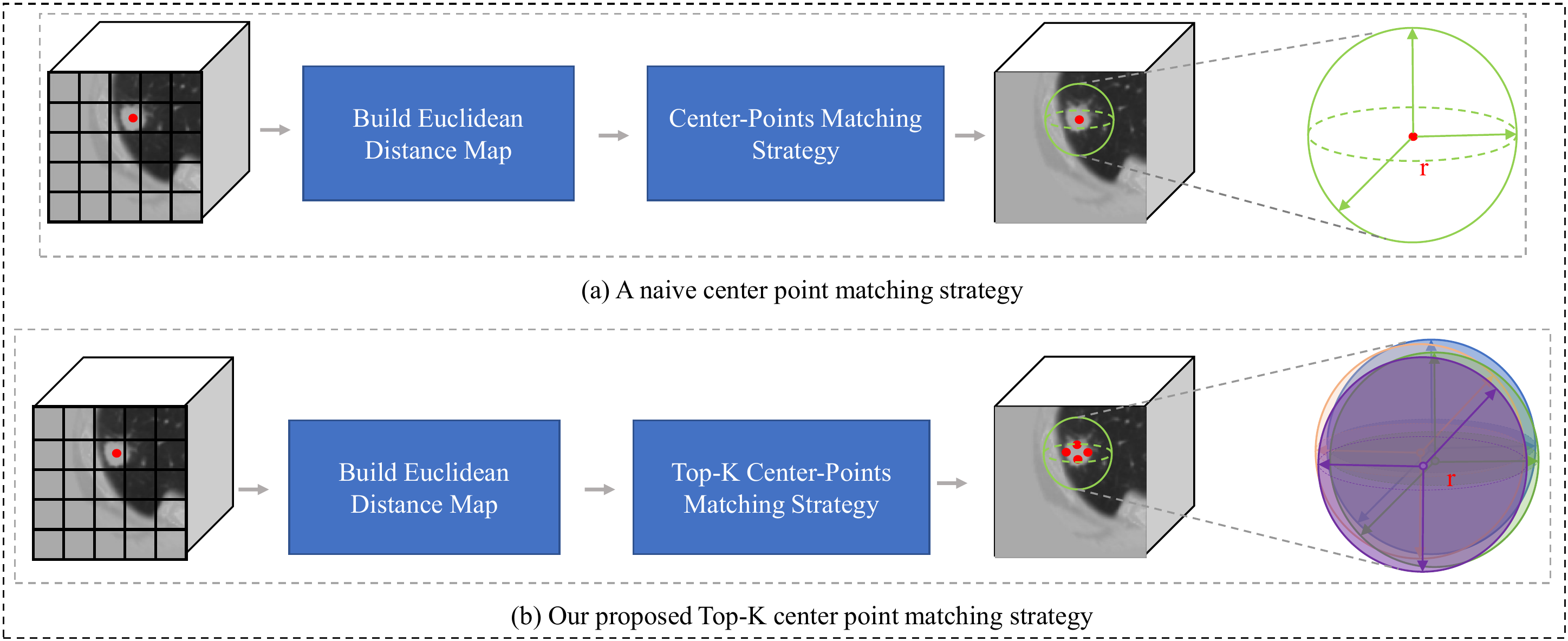}
    \caption{Illustration of center-point matching used for training. (a) is a simple center point matching strategy, where a nodule is only associated with a single centroid. (b) is our proposed top-\textit{K} center points matching strategy, where we find \textit{K} points around the centroid of the nodule.}
\label{fig:point_matching}
\end{figure*}
\begin{figure}
    \centering
    \includegraphics[width=0.48\textwidth]{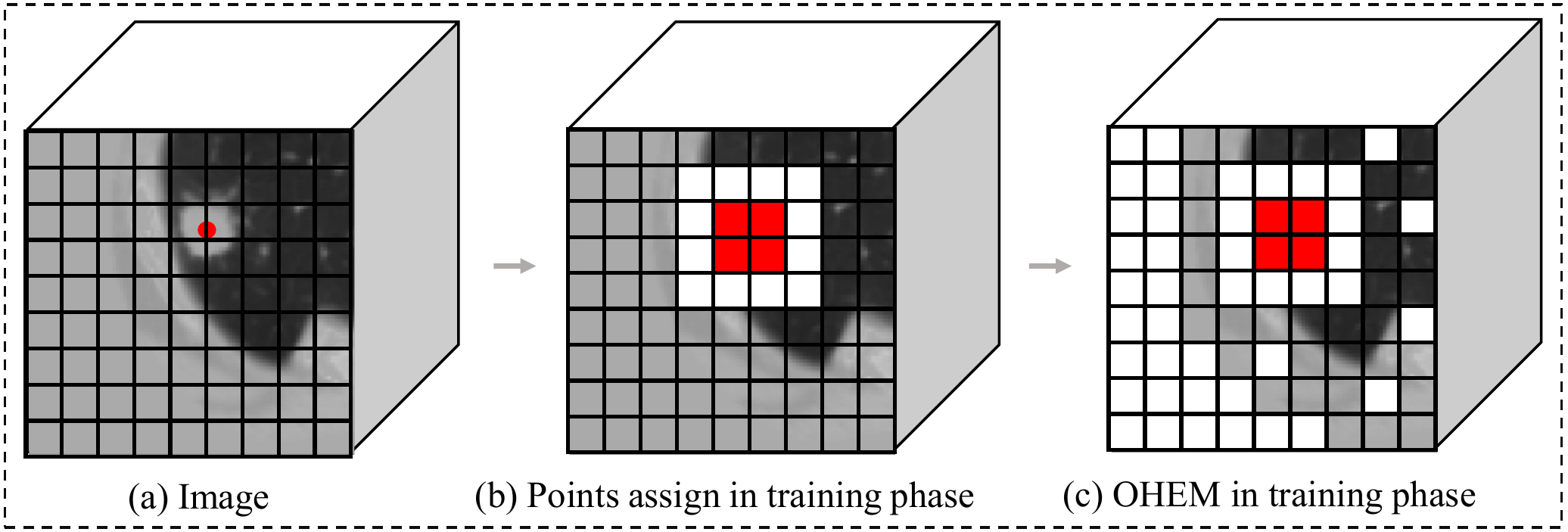}
    \caption{The points assigning process in the training phase. Red, white and the others grids mean positive, ignored and negative points respectively.}
    \label{fig:points_ng_pos}
\end{figure}

\section{Method}
The overall structure of the proposed SCPM-Net is illustrated in Figure~\ref{fig:framework}. It consists of a feature extractor that takes advantage of multi-level spatial coordinate maps and Squeeze-and-Excitation (SE)-based channel attention for better feature extraction, and a head that predicts the existence of a nodule and its radius and offset simultaneously at two resolution levels. To assist the training process, we propose a center points matching strategy to predict \textit{K} points that are nearest to the center of a nodule. Inspired by clinical guidance~\citep{macmahon2017guidelines}, we use a novel sphere representation of nodules to improve the detection accuracy. We represent the network's 3D input as $ I \in \mathbb{R}^{D \times H \times W}$, where \textit{D}, \textit{H} and \textit{W} denote depth, height and width of the input image, respectively. The predicted feature map with the size of $\frac{D}{R} \times \frac{H}{R} \times \frac{W}{R} \times C$ can be obtained by forward process of the SCPM-Net, where \textit{R} denotes down-sampling ratio and \textit{C} denotes channel number. Let $M_C$ represent the center point map with one channel indicating the probability of each pixel (i.e., point) being close to the center of a nodule, $M_R$ represent the radius map with one channel where each element gives the radius of the nodule centered at that pixel, and $M_O$ represent the offset map with three channels indicating offsets of the center point in the directions of x, y, and z, respectively.
\subsection{Architecture of the CPM-Net}
\begin{figure*}
    \centering
    \includegraphics[width=1.0\textwidth]{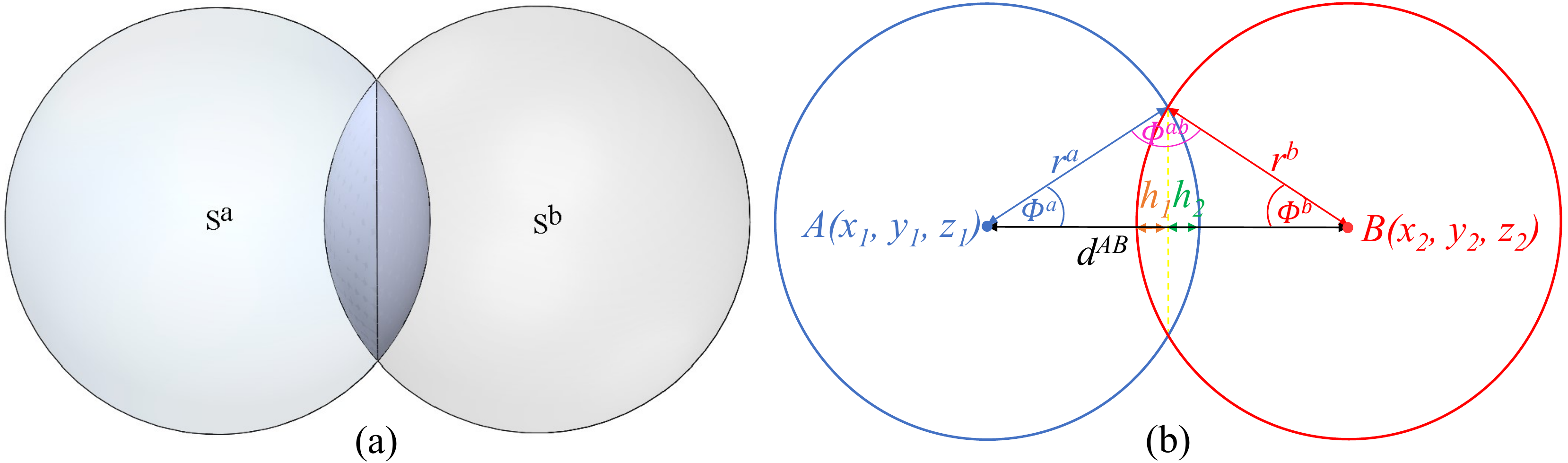}
    \caption{Illustration of SIoU calculation. (a) An example of intersection between two spheres; (b) A projection view of the intersection area between two spheres. }
    \label{fig:2d_sphere}
\end{figure*}
The architecture of our detection backbone follows an encoder-decoder flowchart as shown in the left of Figure~\ref{fig:framework}. It has a series of convolutional layers to learn 3D patterns. Each convolution uses 3$\times$3$\times$3 kernels with 1 as stride, and each down-sampling and up-sampling is implemented by max-pooling and deconvolution, respectively. Skip connections are used to link low-level features and high-level features. Each of the last two decoder blocks in our CPM-Net obtains three prediction maps, i.e. center point map $M_C$, radius map $M_R$, and offset map $M_O$, respectively. To be clear, the center point map ($M_C$) is a pixel-wise binary classification map, where each element gives the probability of the corresponding pixel being a center point. Following~\citep{zhao2019object}, an offset map ($M_O$) is used to represent the offset from the current point to the real centroid of a nodule and a radius map ($M_R$) is used to represent the predicted radii of the bounding spheres. In particular, we introduce multi-path normalized coordinate map fusion blocks to generate attentive features for being aware of spatial position. These coordinate maps are generated by using three channels to encode the spatial coordinates along x, y, and z directions, respectively. We obtain a pyramid of the coordinate map by downscaling it into four resolution levels, with the size of $48^3$, $24^3$, $12^3$, and $6^3$ respectively. At each resolution level of the pyramid, we concatenate the coordinate map with the image features. Following~\citep{hu2018squeeze}, SE-blocks are adopted to learn spatial and channel-wise attentions to enhance more discriminative features. In the last block of the encoder, we utilize a dilated fusion block, consisting of three dilated convolutions with dilation rates of 1, 2, 3 respectively. In the training phase, we use a proposed center-points matching procedure as mentioned in Sec.~\ref{method:cpm} and a combination loss as described in Sec.~\ref{sec:ohem}. In the testing phase, a bounding sphere of detection result can be constructed by the position of predicted centroids, radius, and offset as clarified in Sec.~\ref{sec:imp}.

\subsection{Center-Points Matching}\label{method:cpm}Anchor-based methods assign positive and negative anchors according to the threshold of overlap between anchor boxes and ground-truth for training. In contrast, without any anchor prior, a novel center-points matching strategy for training is proposed in this work to assign positive, negative, and ignored labels to make the network predict whether a point belongs to center points of lung nodules. In short, this center-points matching process is to generate classification labels in the training phase on the fly. Figure~\ref{fig:point_matching} shows two different point matching strategies in the training phase: (a) a simple center point matching strategy, where the nodule is only associated with one centroid. (b) our proposed top-\textit{K} points matching strategy, which uses \textit{K} points near the centroid to represent a nodule. 
\par The whole strategy can be divided into four steps: 1) calculating the distance between each point to the ground truth centroid and using the distance map as a prior to assigning positive, negative, and ignored points; 2) selecting top-\textit{K} points that are nearest to a centroid as the positive point set $\mathcal{P}$; 3) to reduce the false positive, a certain amount of points around positive points are put into the set of ignored points $\mathcal{I}$. The remaining points are treated as a temporary negative points-set $\mathcal{N}$; 4) using online hard example mining (OHEM~\citep{shrivastava2016training}, detailed in Sec.~\ref{sec:ohem}) to further sample hard negative points as point set $\mathcal{N^0} \subset  \mathcal{N}$, while points in $\mathcal{N} - \mathcal{N^0}$ are reset to the ignored points, and therefore we can obtain an ignored point set $\mathcal{I^0}$ = $\mathcal{I} + \mathcal{N} - \mathcal{N^0}$. The points assigning process in the training phase is shown in Figure~\ref{fig:points_ng_pos}. It is worth mentioning that this matching strategy can be used to train any center points-based anchor-free detector. But in this paper, we pay more attention to training a detector to predict the bounding sphere. We loop all the annotations in the image using the above steps. Note that the top-\textit{K} positive points are selected according to the lowest distance values sorted in the distance map, and ignored points do not participate in training.    

% Leveraging each ground-truth, the distance between two points can be calculated by Equation~\ref{equ:euq_dist},
% \begin{equation}\label{equ:euq_dist}
% \begin{aligned}
%     dist=\left(x_{d, h, w}-\frac{x^{\prime}}{R}\right)^{2}
%     +\left(y_{d, h, w}-\frac{y^{\prime}}{R}\right)^{2}
%     +\left(z_{d, h, w}-\frac{z^{\prime}}{R}\right)^{2}\right
% \end{aligned}
% \end{equation}
% here, $(z^{\prime}, y^{\prime}, x^{\prime})$ denotes the coordinate of the center point of ground truth. 
% \begin{figure}
%     \centering
%     \includegraphics[width=0.48\textwidth]{3d_sphere.pdf}
%     \caption{An example of intersection between two spheres.}
%     \label{fig:3d_sphere}
% \end{figure}
\subsection{Sphere Representation and Sphere-Based Loss Functions}
\subsubsection{Similarity based on Sphere Representation}Due to the ellipsoid-like shape, lung nodules can be better represented by the bounding sphere than the bounding box~\citep{sladoje2005measurements}. In clinical pulmonary nodule diagnosis, it is a common practice that using the coordinate and diameter to represent the nodule in 3D space~\citep{setio2017validation, macmahon2017guidelines}. Based on these observations, we use the sphere representation for nodules in 3D space. First, we introduce a sphere-based intersection-over-union (SIoU) to measure the similarity between a predicted nodule and the ground truth nodule. The SIoU similarity is defined as:

\begin{equation}\label{equ:siou}
    SIoU = \frac{|S^{a} \cap S^{b}|}{|S^{a} \cup S^{b}|}
\end{equation}where $S^{a}$ and $S^{b}$ denote the predicted bounding sphere and the ground truth bounding sphere respectively as shown in Figure~\ref{fig:2d_sphere} (a). To better formulate this method, we use a projection view to illustrate calculation details in Figure~\ref{fig:2d_sphere} (b). The centroids of sphere $A$ and sphere $B$ are defined as $A(x_1, y_1, z_1)$ and $B(x_2, y_2, z_2)$ in image space respectively. $r^a$ and $r^b$ denote the radius of sphere $A$ and sphere $B$ respectively. The distance of between the central coordinates of sphere $A$ and $B$ is defined as:
\begin{equation}\label{eq:dis}
    d^{AB} = ||A-B||_2
\end{equation}When $S^a$ and $S^b$ are intersected, i.e., $r^a + r^b > d^{AB}$, let $\phi^a$ and $\phi^b$ denote the central angles of the two spheres, respectively, and let $\phi^{ab}$ denote the intersection angle. $h_1$ and $h_2$ represent the distance between the intersected chord to the arcs, respectively, as shown in Figure~\ref{fig:2d_sphere}(b). These values and the intersection between A and B can be obtained by:
\begin{equation}
    cos(\phi^a) = \frac{(r^a)^2 + (d^{AB})^2 - (r^b)^2}{2  r^a  d^{AB}}
\end{equation}

\begin{equation}
    cos(\phi^b) = \frac{(r^b)^2 + (d^{AB})^2 - (r^a)^2}{2  r^b  d^{AB}}
\end{equation}

\begin{equation}\label{equ:cos}
    cos(\phi^{ab}) = \frac{(r^b)^2 + (r^a)^2 - (d^{AB})^2}{2  r^a  r^{b}}
\end{equation}

\begin{equation}
    h_1 = r^b (1 - cos(\phi^b))
\end{equation}

\begin{equation}
    h_2 = r^a  (1 - cos(\phi^a))
\end{equation}

\begin{equation}\label{eq:interaction}
    |S^{a} \cap S^{b}| = \pi  r^a  h_2^2 -
    \frac{\pi  h_2^3}{3} + \pi  r^b  h_1^2 -
    \frac{\pi h_1^3}{3}
\end{equation}When $S^a$ and $S^b$ are not intersected, i.e., $r^a + r^b <= d^{AB}$, $| S^a \cap S^b |$ is 0. The union between $S^a$ and $S^b$ is defined as:
\begin{equation}\label{eq:union}
|S^{a} \cup S^{b}| = \frac{4\pi((r^a)^3 + (r^b)^3)}{3} - |S^{a} \cap S^{b}| 
\end{equation}Based on these sub-modules, we can calculate the SIoU similarity of two spheres using Eq.~\eqref{equ:siou}.

\subsubsection{Sphere-Based Intersection-over-Union Loss Function}The effectiveness of IoU-based loss function
has been well proven for 2D detection tasks~\citep{rezatofighi2019generalized, zhou2019iou, zheng2020distance}. However, there are few works that focus on designing loss function for 3D object detection, especially in 3D medical images. Based on our proposed sphere representation, we introduce a Sphere Intersection-over-Union (SIoU) loss for 3D nodule detector training which is inspired by~\citep{zhou2019iou, rezatofighi2019generalized, zheng2020distance}:
\begin{equation}
    L_{SIoU} = 1 - SIoU
\end{equation}where $SIoU$ is defined in Eq.~\eqref{equ:siou}. Similarly with traditional Intersection over Union loss function $L_{IoU}$~\citep{zhou2019iou}, $L_{SIoU}$ only works when the predicted sphere and the ground truth have overlap, and would not provide any moving gradient for non-overlapping cases. To deal with the cases where $S^a$ and $S^b$ are not intersected, we additionally use distance and radius ratio ($R_{DR}$) for optimization.
\begin{equation}\label{eq:dr}
    R_{DR} = \frac{d^{AB}}{d^{AB} + r^a + r^b}
\end{equation}Then, we integrate the $R_{DR}$ into $L_{SIoU}$:
\begin{equation}
    L_{SDIoU} = 1.0 + R_{DR} - SIoU
\end{equation}In addition, we found that the angle of intersection $\phi^{ab}$ between the two spheres can also be used to better describe the regression of the sphere. We therefore define a angle score $\eta$ as follows:
\begin{equation}\label{eq:eta}
    \eta =\left\{\begin{array}{l}0\text{,~if~} d^{AB}>(r^a + r^b)\\ 
    \frac{\arccos{(cos(\phi^{ab}))}}{\pi}
    \text{,~otherwise}
\end{array}\right.
\end{equation}where $cos(\phi^{ab})$ have been calculated in Eq.~\ref{equ:cos}. Finally, we integrated all geometric measures (sphere iou, distance and radius ratio, angle of intersection) into a unified loss, which is referred to as $L_{SIoU++}$:
\begin{equation}\label{eq:siou++}
L_{SIoU++}=\left\{\begin{array}{l}R_{DR}\text{,~if~} d^{AB}>(r^a + r^b)\\ 1.0 + R_{DR} - SIoU + \eta \text{,~otherwise}
\end{array}\right.
\end{equation}It can be observed that the sphere representation considers many geometric measures, i.e., overlap area ($S^a \cap S^b$), central point distance ($d^{AB}$) and angle of intersection ($\eta$), which have been ignored in the other representations. The pseudo-code for $L_{SIoU++}$ computation with two spheres is given in Alg.~\eqref{alg:alg1}. The properties of $L_{SIoU++}$ function can be summarized as follows: 
\begin{itemize}
\item[1)]Like the genral $L_{IoU}$, our $L_{SIoU++}$ is  invariant to the scale of regression problem.
\item[2)]$L_{SIoU++}$ can alleviate the gradient vanishing problem when there is not overlap between prediction and the ground truth.
\item[3)]$L_{SIoU++}$ converges much faster than $L_{IoU}$, since $L_{SIoU++}$ minimizes the overlap, distance, radius ratio and angle of intersection directly.
\end{itemize}

% \begin{algorithm}[t]
% \normalem
% \caption{$L_{SIoU++}$ Function for 3D Detector Training.}
% % \LinesNumbered
% \KwIn{Central coordinates $A(x_1, y_1, z_1)$ and $B(x_2, y_2, z_2)$ and radii $r^a$ and $r^b$ of two spheres in image space respectively.}
% \KwOut{$L_{SIoU++}$;}
% \State $d^{AB} = ||A-B||_2$;\\
% \State $R_{DR} = \frac{d^{AB}}{d^{AB} + r^a + r^b}$;\\
% \If{($r^a$ + $r^b$) \textgreater \ $d^{AB}$}
%         {
%             \If{($r^a$ + $d^{AB}$) \textless \ $r^b$}
%             {
%                 $SIoU = (\frac{r^a}{r^b})^3$;
%             }
%             \ElseIf
%             {($r^b$ + $d^{AB}$) \textless $r^a$}
%             {
%                 $SIoU = (\frac{r^b}{r^a})^3$;
%             }
%             \Else{
%             $SIoU = \frac{|S^{a} \cap S^{b}|}{|S^{a} \cup S^{b}|}$;
%             }}
%         \Else
%         {
%             $SIoU = 0$;
% }\\
% \State $L_{SIoU++} = 1.0 + R_{DR} - SIoU + \eta$;\\ 
% \Return $L_{SIoU++;}$
% \label{alg:alg1}
% \end{algorithm}

\begin{algorithm}[htp]
\normalem
	\caption{Pseudocode of $L_{SIoU++}$ for 3D Lung Nodules Detector Training.}
	\label{alg:alg1}
	\KwIn{Central coordinates $A(x_1, y_1, z_1)$ and $B(x_2, y_2, z_2)$ and radii $r^a$ and $r^b$ of two spheres in image space respectively.}
	\KwOut{$L_{SIoU++}$;}
	\textbf{Initialization:} $d^{AB} = ||A-B||_2$; $R_{DR} = \frac{d^{AB}}{d^{AB} + r^a + r^b}$;
	\BlankLine
	\If{($r^a$ + $r^b$) \textgreater \ $d^{AB}$}
	    {
	    \textbf{Calculate} $|S^{a} \cap S^{b}|$ according to Eq.~\eqref{eq:interaction};\\
	    \textbf{Calculate} $|S^{a} \cup S^{b}|$ according to Eq.~\eqref{eq:union};\\
	    \textbf{Calculate} the $SIoU$ according to Eq.~\eqref{equ:siou};\\
	    \textbf{Calculate} the angle score $\eta$ using Eq.~\eqref{eq:eta};\\
	    \textbf{Calculate} $L_{SIoU++}$ = 1.0 + $R_{DR}$ - $SIoU$ + $\eta$;}
	\Else{
	    $L_{SIoU++}$ = $R_{DR}$;
	}
    \Return $L_{SIoU++}$;
\end{algorithm}

\subsection{The Overall Loss Function}\label{sec:ohem} The imbalance of positive and negative samples in 3D point classification is more severe than that in 2D point classification. A hybrid method of Online Hard Example Mining (OHEM)~\citep{shrivastava2016training} and re-focal loss~\citep{lin2017focal} are used in our SCPM-Net to deal with such a huge imbalance. In OHEM, we select some hardest negative point samples for training. Specifically, all the negative points are sorted in the descending order of the center point classification loss, and we select the first $N$ points of them after the sorting. $N$ is set to $n$ times of $M$ if $M > 0$ and 100 otherwise, where $M$ is the number of positive points in the input image and $n$ is the ratio between negatives and positives samples. In the end, negative points those are not selected are reset as ignored points. A re-focal loss is shown as Eq.~\eqref{equ:rfocal}, which can improve the sensitivity and further balance the gradient of the positives and the negatives.
 \begin{equation}\label{equ:rfocal}
 \begin{aligned}
    L_{cls}=\sum_{j=0}^{J}-w_{j}\alpha (1 - p_{j}) & ^{\gamma} \log (p_j)
\end{aligned}
\end{equation}The weight of re-focal loss is defined as: \begin{equation}
    w_{j} = \left\{\begin{array}{l}1 \text{,~~if}\ {j} \in \mathcal{P}\ \text { and } \ p_{j}>{t} \text{~or~}\; {j} \in \mathcal{N} \\0 \text{,~~if} \ {j} \in \mathcal{I} \\ {w} \text{,~~if} \ {j} \in \mathcal{P} \text { and } \; p_{j}<{t}\end{array}\right.
\end{equation}
where $p_{j}$ denotes the probability of point \textit{j} being the centroid of a nodule. $J=\frac{D}{R} \times \frac{H}{R} \times \frac{W}{R}$
is the total number of points. \textit{t} is a threshold to filter unqualified points, and \textit{w} is a weight to balance the re-focal loss. A smooth $L_1$ loss~\citep{ren2015faster} is also used to regress the normalized radius of bounding sphere:
\begin{equation}\label{equ:size}
% \begin{align}
   L_{radius}\left(r, r^{*}\right)=
%   \\ \sum_{r \in\{d, h, w\}}
  \\ \left\{\begin{array}{ll}\frac{0.5 (r -  r^{*})^{2}}{\beta}\text {,~~if }|r -  r^{*}|<\beta \\ |r -  r^{*}|\text{,~~otherwise }\end{array}\right. 
% \end{align}
\end{equation}where $r$ denotes predicted radius value, and $r^{*}$ is the ground truth radius value. To obtain more accurate locations of small objects, we use $L2$ loss to regress the offset between the center points ground truth and the predicted positive points.
\begin{equation}
   L_{offset}\left(f, f^{*}\right)=||f -  f^{*}||_2
\end{equation}where $f$ and $f^*$ are 3D vectors and denote the predicted and ground truth offsets, respectively. The total loss is expressed by the following equation:
\begin{equation}\label{equ:total_loss}
\begin{aligned}
  L_{total} = L_{cls} + \alpha(L_{radius} + L_{offset} + \lambda_{s} L_{SIoU++})
\end{aligned}
\end{equation}where $L_{radius}$, $L_{offset}$ and $L_{SIoU++}$ are only meaningful for positive points, thus we define $\alpha$ = 1 for points in $\mathcal{P}$ and 0 for points in $\mathcal{I}$ and $\mathcal{N}$. $\lambda_{s}$ represents the weight factor of $L_{SIoU++}$.

\subsection{Implementation Details}\label{sec:imp}
We used PyTorch~\citep{paszke2019pytorch} to implement our SCPM-Net. The training and testing process were done via SenseCare~\citep{duan2020sensecare} platform with one 8-core Intel E5-2650 CPUs, 8 NVIDIA 1080Ti GPUs and 2T memory. During pre-processing, following~\citep{song2020cmpnet,wang2020focalmix,liao2019evaluate}, we resampled all images to the isotropic resolution of 1.0 $\times$ 1.0 $\times$ 1.0 $mm^3$ to reduce the spatial variation among images and normalized them to zero mean and unit variance. The coordinate map was generated based on the entire image and re-scaled to [0-1] and saved into memory, before the training stage. In the training phase, we randomly cropped an image patch and coordinate map with a size of 96 $\times$ 96 $\times$ 96 as the input of SCPM-Net. For each model, we used 170 epochs in total with stochastic gradient descent optimization, momentum  0.9, and weight decay  $10^{-4}$. The batch size was  24. The initial learning rate was $10^{-4}$ in the first 20 epochs for warm up training, and then set as $10^{-2}$, $10^{-3}$ and $10^{-4}$ after 20, 80 and 150 epochs, respectively.
\par In the inference phase, we used sliding window with a stride of $24 \times 24 \times 24$ to obtain the whole scan's prediction. Then, we picked up top-\textit{n} candidate predicted points $[{p}_{0}, {p}_{1}, ..., {p}_{n-1}]$ in a classification map with a size of $\textit{D}\times\textit{H}\times\textit{W}\times1$ by sorting the probabilities. Each predicted candidate point has a probability ${p}_{j}$ in a 3D integer location $\mathbf{x}_j \in \mathbb{Z}^3$. In the same integer location, we can get an offset prediction $\mathbf{v}_j \in \mathbb{R}^3$ and a radius prediction $r_{j}$, respectively. Then a detected bounding sphere will be located at $\mathbf{x}_j +\mathbf{v}_j$ with a radius of $r_j$. Finally, all the detected spheres will pass a 3D SIoU++ based non-maxima suppression (NMS)~\citep{ren2015faster, zheng2020distance} to filter overlapping spheres. Compared with IoU-based NMS, the SIoU++ based NMS uses SIoU and distance and radius ratio of two spheres as the measures to filter redundant spheres. Parameter setting was \textit{K} = 7, $\lambda_{s}$ = 2, \textit{n} = 100, \textit{t} = 0.9, \textit{w} = 4, $\beta$ = $\frac{1}{9}$, $\alpha$ = 0.375, $\gamma$ = 2 based on a grid search with the validation data.

% In our experiment, K, M, N, $\lambda_{c}$, $\lambda_{r}$, $\lambda_{o}$, $\lambda_{s}$, $\beta$, $\alpha$ and $\gamma$ are set to 7, 100, 10000, 2, 1, 1, 2, $\frac{1}{9}$, 0.75 and 2, respectively.

\section{Experimental Results}
\subsection{Dataset and Evaluation Metrics}
In this work, we validated the proposed framework on the large-scale public challenge dataset of LUNA16~\citep{setio2017validation}, which contains 888 low-dose CT scans with the centroids and diameters of the pulmonary nodules annotated. In the LUNA16 challenge, performances of detection systems are evaluated using the Free-Response Receiver Operating Characteristic (FROC)~\citep{setio2017validation}. The sensitivity and the average number of false positives per scan (FPs/Scan) are used to measure the detector's performance. Following existing works~\citep{cao2020two,song2020cmpnet}, we firstly used the first nine subsets for training and the last subset for validation during ablation study and comparison with several open-source methods in the same settings, and these results are presented in Sec.~\ref{exp:exp1},~\ref{exp:exp2},~\ref{exp:exp3},~\ref{exp:exp4}. Then, we compared the proposed method with state-of-the-art methods using standard 10-fold cross-validation in Sec.~\ref{exp:exp5}, for a fair comparison. We used the official results of these works rather than our re-implemented results.
\subsection{Effect of Center-Points Matching Strategy}\label{exp:exp2}
\subsubsection{Impacts of Different $K$ Values}To investigate the effect of our center-points matching strategy, we set $\lambda_{s}$ in Eq.~\eqref{equ:total_loss} to 0 without any additional efforts, and compared the performance at different \textit{K} values for selecting the top-\textit{K} positive center points of each nodule during training, as mentioned in Sec.\ref{method:cpm}. Figure~\ref{fig:specificy_topk} shows evolution of the detector's performance on the testing subset when \textit{K} changes from 5 to 9. It can be found that when \textit{K} is greater or smaller than 7, the sensitivity goes down significantly. This is because too many positive points during training will bring in false positives while few positive points weaken the model's ability to locate the nodules. Therefore, we set $K=7$ for our center-points matching strategy in the following experiments. 
\subsubsection{Impacts of Different Ratios of Negative and Positives Samples}
\begin{figure}[htbp]
\centering
\begin{minipage}[t]{0.49\textwidth}
\centering
\includegraphics[width=1.0\textwidth]{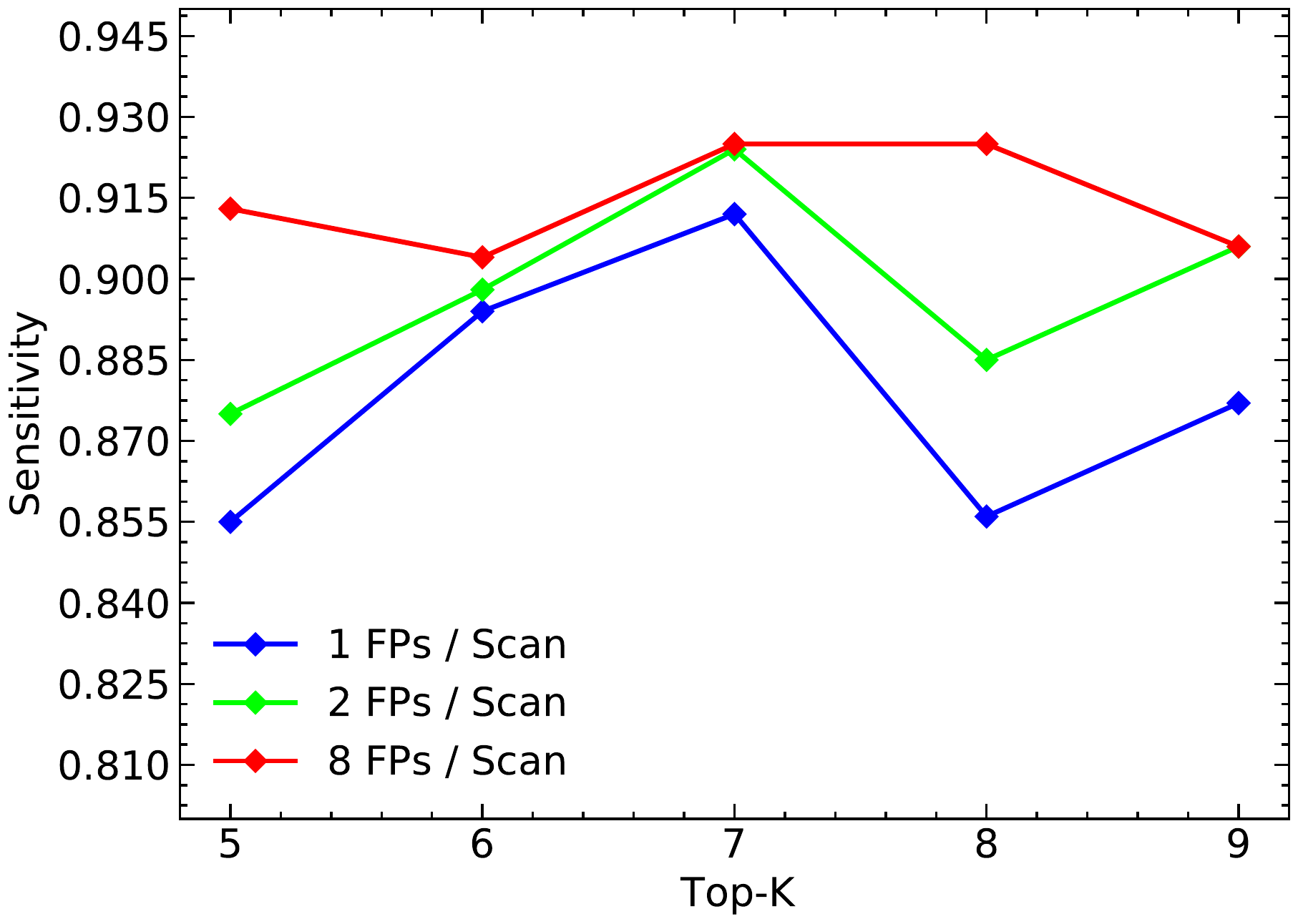}
\caption{The detection performance of CPM-Net with center points matching strategy with different values of \textit{K}.}
\label{fig:specificy_topk}
\end{minipage}
\begin{minipage}[t]{0.49\textwidth}
\centering
\includegraphics[width=1.0\textwidth]{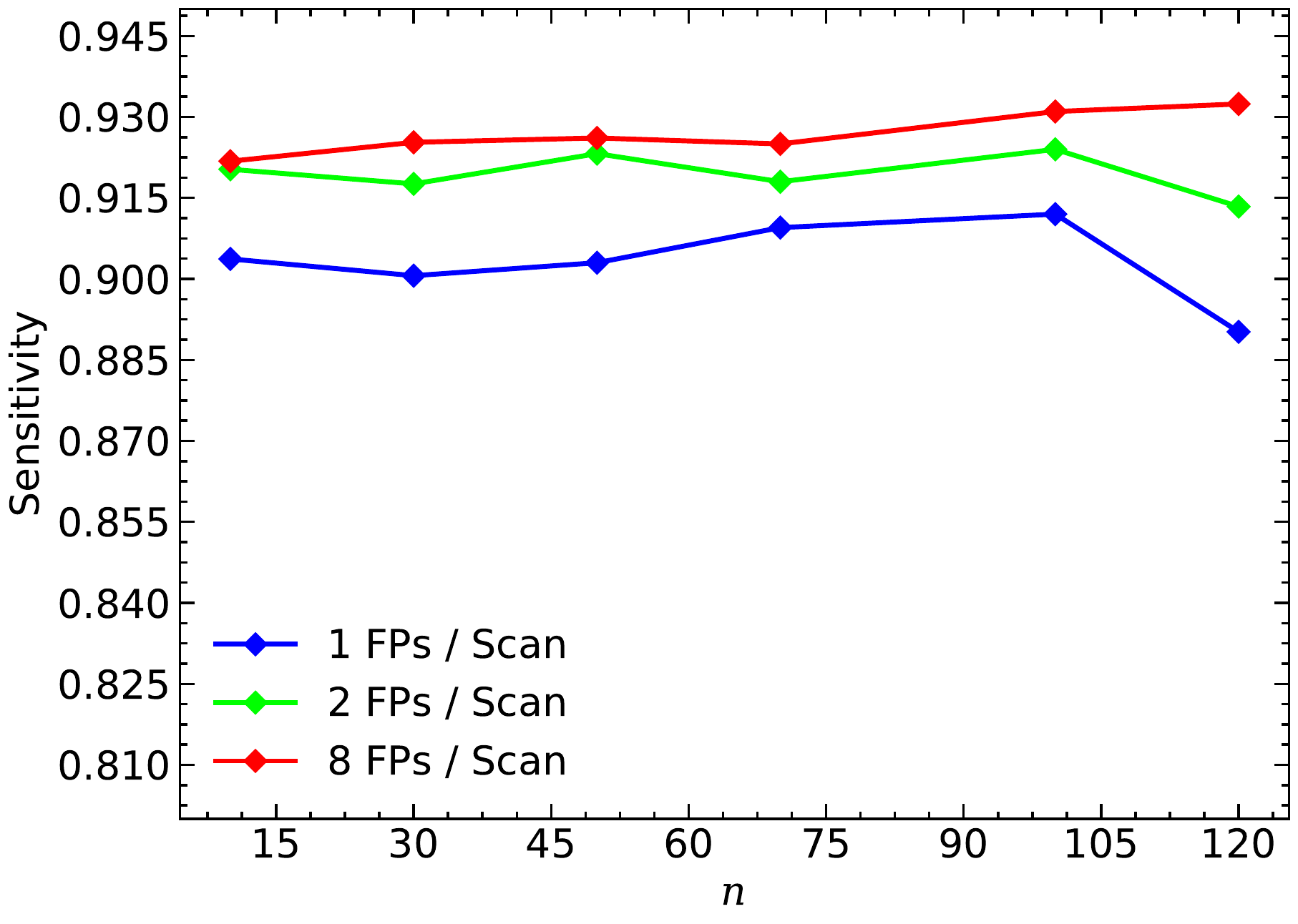}
\caption{The detection performance of CPM-Net with center points matching strategy under different ratios (\textit{n}) between negatives and positives samples.}
\label{fig:specificy_ratio_n}
\end{minipage}
\end{figure}

We further investigate the impacts of detection results under different ratios of negatives and positives samples ($n$), as described in Section~\ref{sec:ohem}. The detection performance at different ratios of negatives and positives samples is
shown in Figure~\ref{fig:specificy_ratio_n}. It can be found that increasing $n$ from 10 to 100 leads to the sensitivity improvement at 1FP/Scan and 2FPs/Scan. When $n$ equals 120 the sensitivity at 1FP/Scan and 2FPs/Scan becomes lower. However, the sensitivity at 8FPs/Scan increases progressively when changing $n$ from 10 to 120. This shows that the ratios of negatives and positives samples affect the detection results in different ways, where a larger ratio may cause more false positives. In the following experiments, we set $n$ to 100, as it can achieve better results at 1FP/Scan and 2FPs/Scan where they are more important in clinical practice.
\subsubsection{Contributions of Different Training Strategies}To measure the contribution of each training strategy used in this work, we also implemented an ablation study based on the proposed detector. The baseline just uses the naive classification loss, radius, and offsets regression losses without OHEM and Focal/Re-Focal Loss strategies. We compared the baseline with: 1) Baseline + OHEM, where the OHEM strategy was used for hard examples mining; 2) Baseline + OHEM + Focal Loss, where a Focal loss was further used to balance the impact of hard and easy samples; 3) Baseline + OHEM + Re-Focal Loss, where a Re-Focal Loss was introduced for better results. Quantitative evaluation results of these strategies are listed in Table~\ref{tab:loss_analysis}. It can be found that using OHEM, Focal Loss, and Re-Focal Loss can achieve better results than the baseline. Compared with the classical Focal Loss~\citep{lin2017focal}, the Re-Focal Loss improves the sensitivity at all Fps/Scan, demonstrating the effectiveness of the proposed Re-Focal Loss. 
\subsection{Analysis of CPM-Net}\label{exp:exp1}The attentive module with Coordinate Attention (CA) and Squeeze-and-Excitation (SE) attention is a crucial component of our CPM-Net. To understand each component's contribution to performance, we perform an ablation study firstly. Following~\citep{song2020cmpnet}, $\lambda_{s}$ in Eq.~\eqref{equ:total_loss} was set as 0 to investigate the effectiveness 
of CPM-Net without any extra effects. In addition, the OHEM and the center-points matching strategy are used to train detectors stably
and efficiently, and \textit{K} and \textit{n} are set to 7, 100 respectively. Then, we trained three models: 1) CPM-Net without both CA and SE attentions; 2) CPM-Net without CA but with SE attention module; 3) CPM-Net with both CA and SE attentions. A quantitative evaluation of this ablation study is listed in Table~\ref{tab:net_analysis}. It shows that adding SE and CA improves the sensitivity by 4.4\% at 1 FPs/Scan, by 3\% at 2 FPs/Scan and by 2.1\% at 8 FPs/Scan, respectively. Then, we investigate the detector performance by predicting the result at different resolution levels of the decoder of CMP-Net. We use CMP-Net$_1$ to denote only predicting at the last block, and use CMP-Net$_2$ to denote predicting at the second to last block. It can be found that CPM-Net$_1$ achieves better results than CPM-Net$_2$. After merging the results of CPM-Net$_1$ and CPM-Net$_2$, the CPM-Net achieves the best results, indicating the two-level outputs can bring performance gain than just a single-level output.
\begin{table}[tb]
\centering
    \scalebox{0.55}{\begin{tabular}{cccc}
 \hline
 \multirow{2}{*}{Module} & \multicolumn{3}{c}{Sensitivity}     \\ \cline{2-4} 
            & FPs/Scan=1      & FPs/Scan=2     & FPs/Scan=8      \\ \hline
Baseline  & 78.7\% &81.2\% & 85.0\% \\ \hline
Baseline + OHEM  & 85.3\% & 88.8\% & 90.5\% \\ \hline
Baseline + OHEM + Focal Loss  & 89.3\% & 90.2\% & 91.5\% \\ \hline
Baseline + OHEM + Re-Focal Loss  & \textbf{91.2\%} & \textbf{92.4\%} &\textbf{92.5\%} \\ \hline
 \end{tabular}}
\caption{Ablation study of CPM-Net with different training strategies. (FPs/Scan: the false positives per scan)}
\label{tab:loss_analysis}
\end{table}

\begin{table}[tb]
\centering
    \scalebox{0.85}{\begin{tabular}{cccc}
 \hline
 \multirow{2}{*}{Module} & \multicolumn{3}{c}{Sensitivity}     \\ \cline{2-4} 
            & FPs/Scan=1      & FPs/Scan=2     & FPs/Scan=8      \\ \hline
 w/o $SE$ and $CA$     & 86.8\%     & 89.4\%     & 90.4\%     \\ \hline
 w/o $CA$    & 90.5\%     & 90.8\%     & 91.5\%     \\ \hline
 CPM-Net           & \textbf{91.2}\%   &  \textbf{92.4}\%   &  92.5\%     \\ \hline
  CPM-Net$_2$  & 90.2\% &91.9\% &92.2\%
  \\ \hline
 CPM-Net$_1$  & 90.8\% & 92.2\% & \textbf{93.2\%} \\ \hline
 \end{tabular}}
\caption{Ablation study of CPM-Net. $SE$ and $CA$ represent squeeze-and-excitation attention and coordinate attention respectively. CPM-Net$_2$ and CPM-Net$_1$ denote the result at the second to last and last decoder blocks respectively, CPM-Net means the merging results of CPM-Net$_2$ and CPM-Net$_1$. (FPs/Scan: the false positives per scan)}
\label{tab:net_analysis}
\end{table}

\begin{table}
\centering
\scalebox{0.8}{\begin{tabular}{cccc}
 \hline
 \multirow{2}{*}{Method} & \multicolumn{3}{c}{Sensitivity}     \\ \cline{2-4} 
            & FPs/Scan=1      & FPs/Scan=2     & FPs/Scan=8 
            \\ \hline
% % # without ohem
 CPM-Net   & 91.2\% & 92.4\% & 92.5\% \\ \hline
 CPM-Net + $L_{IoU}$     & 90.4\%     & 91.3\%     & 94.2\%    
 \\ \hline
CPM-Net + $L_{SIoU}$      & 90.0\%     & 92.4\%     & 93.5\%     \\ \hline
CPM-Net + $L_{SIoU}$ + $L_{R_{RD}}$    & 90.5\%     & 92.1\%     & 93.0\%     \\ \hline
CPM-Net + $L_{SIoU++}$ (SCPM-Net)       & \textbf{92.3}\%   &  \textbf{92.7}\%   &  \textbf{94.8}\%     \\ \hline
 \end{tabular}}
\caption{Ablation study of $L_{SIoU++}$ function. The $L_{SIoU++}$ consists of sphere IoU term (SIoU), angle of intersection and distance-radius-ratio term ($R_{RD}$). (FPs/Scan: the false positives per scan)}
\label{tab:siou_ablation}
\end{table}

\subsection{Effects of Sphere-Based Intersection-over-Union Loss}\label{exp:exp3}
\begin{figure}[t]
    \centering
    \includegraphics[width=0.48\textwidth]{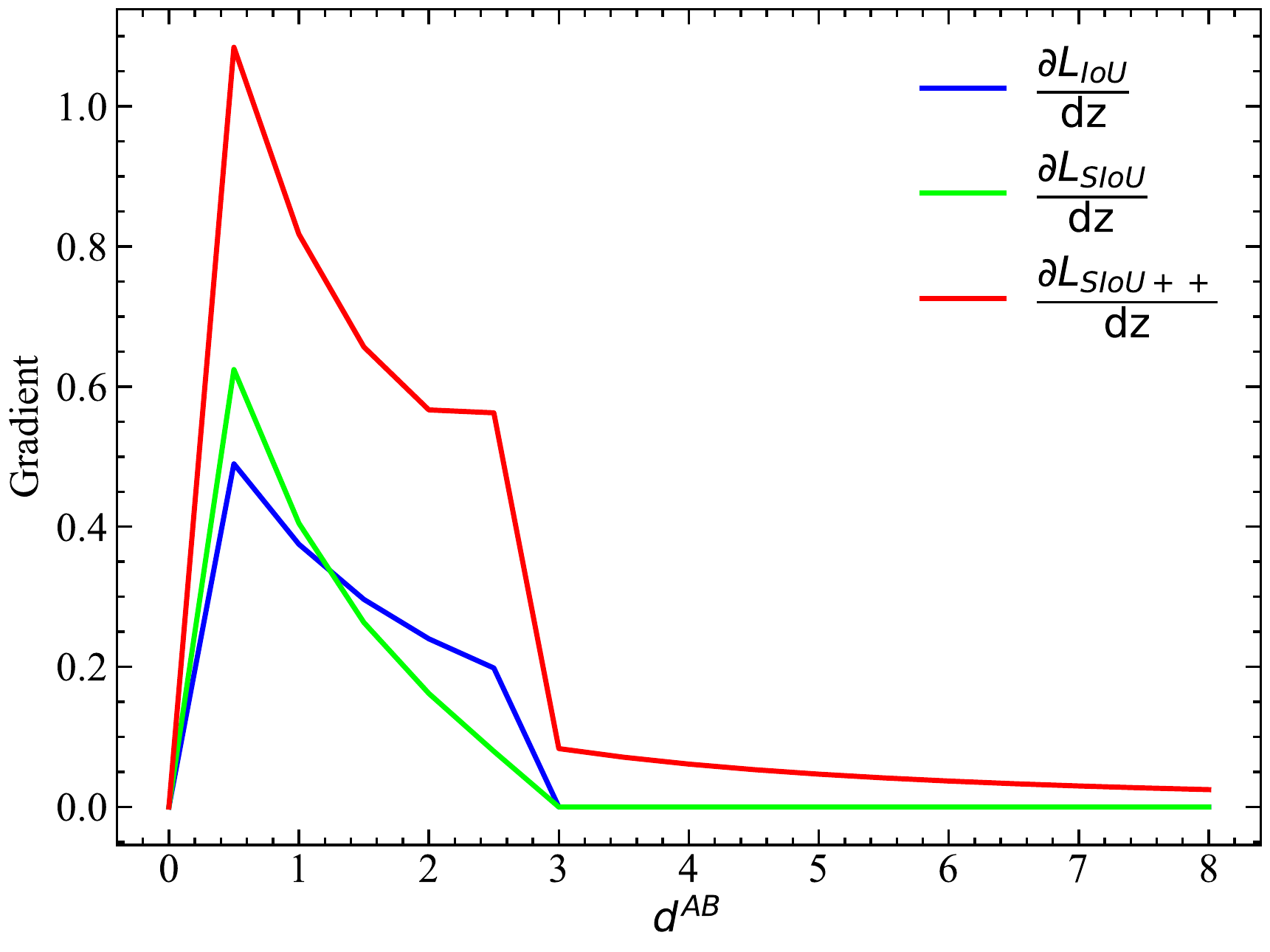}
    \caption{Comparison of the gradients of different loss functions with respect to $z$ when $S^a$ gradually moves to $S^b$ along the $z$-axis ($d^{AB}$ changes from 8 to 0).}
    \label{fig:gradient_curve}
\end{figure}

We first analyzed the difference of gradient between $L_{IoU}$, $L_{SIoU}$ and $L_{SIoU++}$ by simulating the procedure of sphere regression using gradient descent algorithm. To better formulate this study, we suppose that the predicted sphere $S^a$ is at (0, 0, -8) and and the ground truth sphere $S^b$ is at the origin of 3D coordinate. Assume both of them have a radius of 1.5. Then, we calculate the partial derivatives: $\frac{\partial L_{IoU}}{\mathrm{d} z}$, $\frac{\partial L_{SIoU}}{\mathrm{d} z}$, $\frac{\partial L_{SIoU++}}{\mathrm{d} z}$ when $S^a$ gradually moves to  $S^b$ along the $z$-axis. Figure~\ref{fig:gradient_curve} shows evolution of these gradient values with different $d^{AB}$. Compared with $L_{IoU}$ and $L_{SIoU}$, $L_{SIoU++}$ can provide optimization directions for non-overlapping cases (i.e., $d^{AB} > 3$). In addition, $L_{SIoU++}$ has higher gradient values than $L_{IoU}$ and $L_{SIoU}$, since $L_{SIoU++}$ is a joint optimization loss which considers more geometric measures (overlap, distance, radius ratio and angle of intersection).
\par
\begin{figure}
    \centering
    \includegraphics[width=0.48\textwidth]{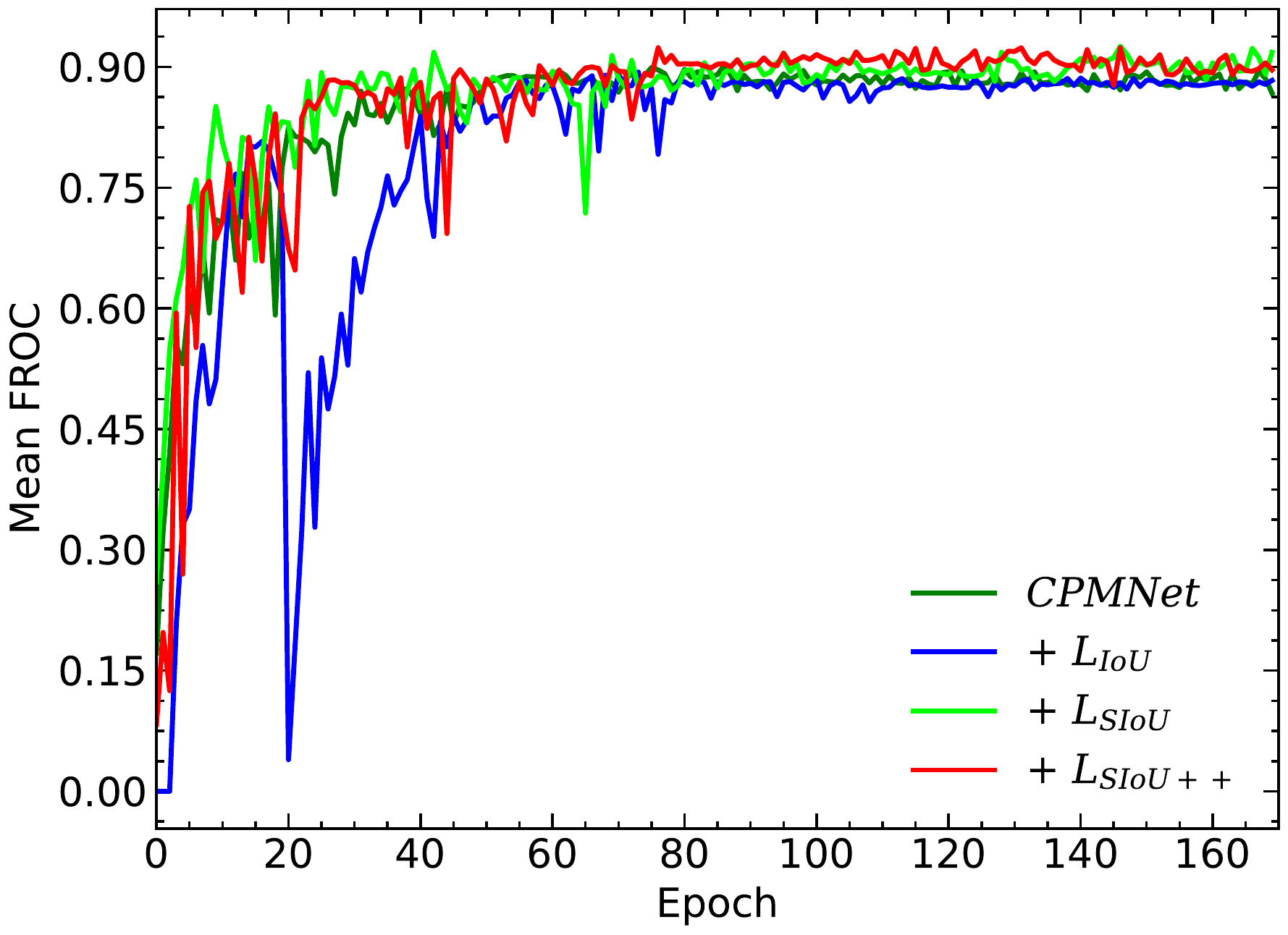}
    \caption{The evolution curve of mean FROC on the validation set. The green, blue, lime, and red curves show the performance of the CPM-Net and the CPM-Net with the $L_{IoU}$, $L_{SIoU}$ and  $L_{SIoU++}$ respectively.}
    \label{fig:convergence_curve}
\end{figure}

 We further analyzed the ability of convergence of the baseline with or without using $L_{SIoU++}$, $L_{SIoU}$ and $L_{IoU}$. In this study, we used the powerful CPM-Net as a baseline and then further combined $L_{SIoU++}$, $L_{SIoU}$ and $L_{IoU}$ with the baseline to train the detector. The evolution of mean FROC curve on the validation set is presented in Figure~\ref{fig:convergence_curve}. It shows that $L_{SIoU++}$ accelerates the convergence of the baseline and leads to better performance than $L_{SIoU}$ and $L_{IoU}$. Note that there exists a sudden drop of mean FROC at the 20th epoch, which is because we used learning rate warm up strategy~\citep{gotmare2018closer} to boost the network's performance. Figure~\ref{fig:convergence_curve} also shows that $L_{SIoU++}$ is more stable than $L_{IoU}$ and $L_{SIoU}$ when the learning rate changed.

\par Finally, we investigated the contribution of each submodule in $L_{SIoU++}$ function and also compared $L_{SIoU++}$ with $L_{IoU}$ with the same experimental setting. A quantitative evaluation of these methods is listed in Table~\ref{tab:siou_ablation}. It shows that using $L_{IoU}$ to train a strong network (CPM-Net) leads to worse results in 1 and 2 FPs/Scan than using $L_{SIoU}$ with $R_{RD}$ and the baseline. In contrast, $L_{SIoU++}$ has the ability to boost the network's performance. Meanwhile, we found that the distance term $R_{RD}$ and angle of intersection term $\eta$ impact the results in different ways but the combination of these terms leads to better than baseline. It further demonstrated that more geometric measures can boost the model to capture more geometric information rather than just focus on reducing regression error. In addition, we also investigated the computational cost of these geometric measure metrics in the training stage in the same settings, we found that all of them require very close training time (2.4s each iteration), which indicates that $L_{SIoU++}$ can boost detection performance without computational cost increasing.
\begin{figure}[t]
    \centering
    \includegraphics[width=0.5\textwidth]{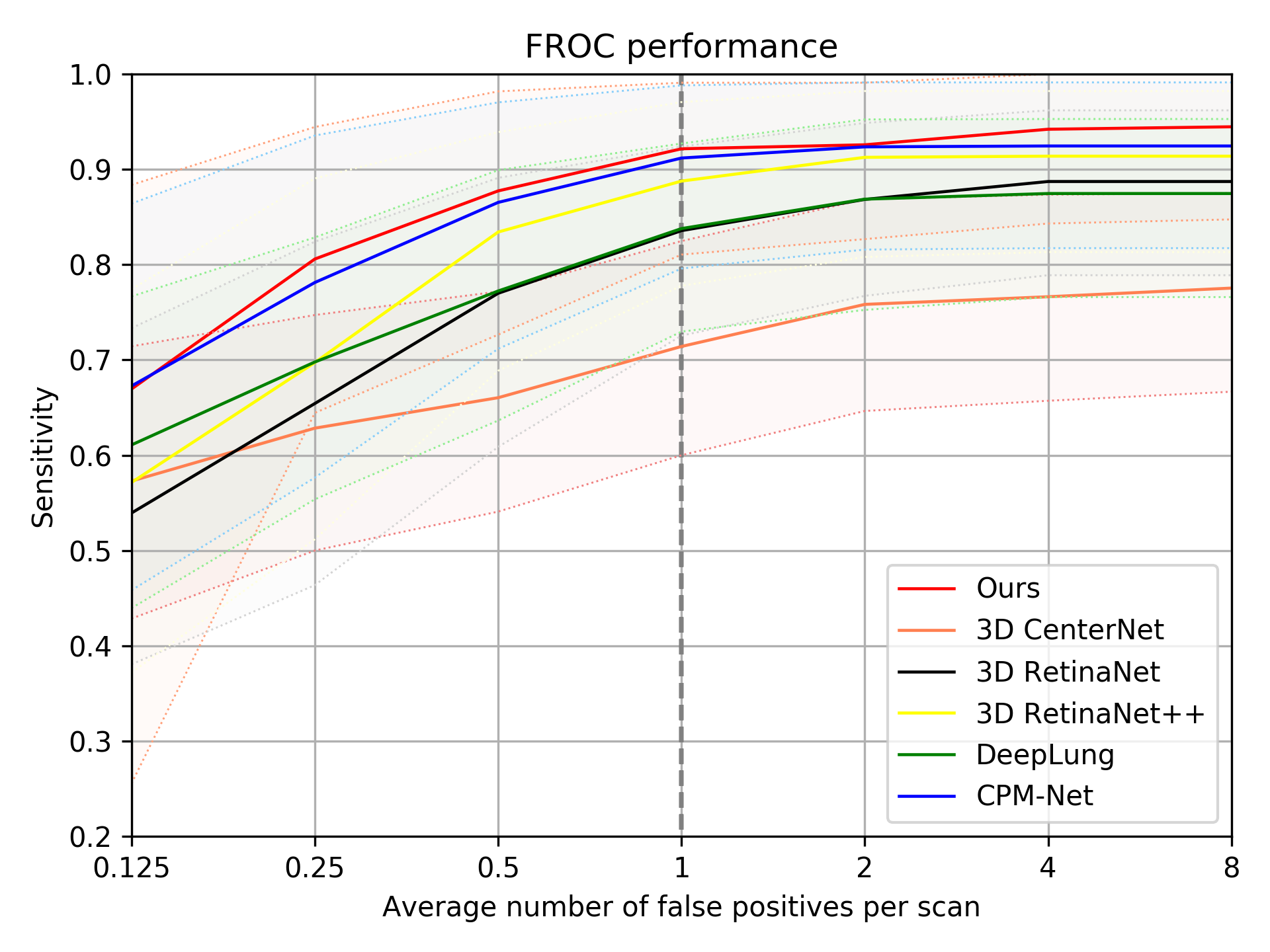}
    \caption{Comparison of FROC curves using different network configurations, with shaded areas presenting the  95\% confidence interval.}
    \label{fig:froc}
\end{figure}

\subsection{Comparison with Existing Methods}
\subsubsection{Comparison with Different Detectors}\label{exp:exp4} 
\begin{figure}[t]
    \centering
    \includegraphics[width=0.48\textwidth]{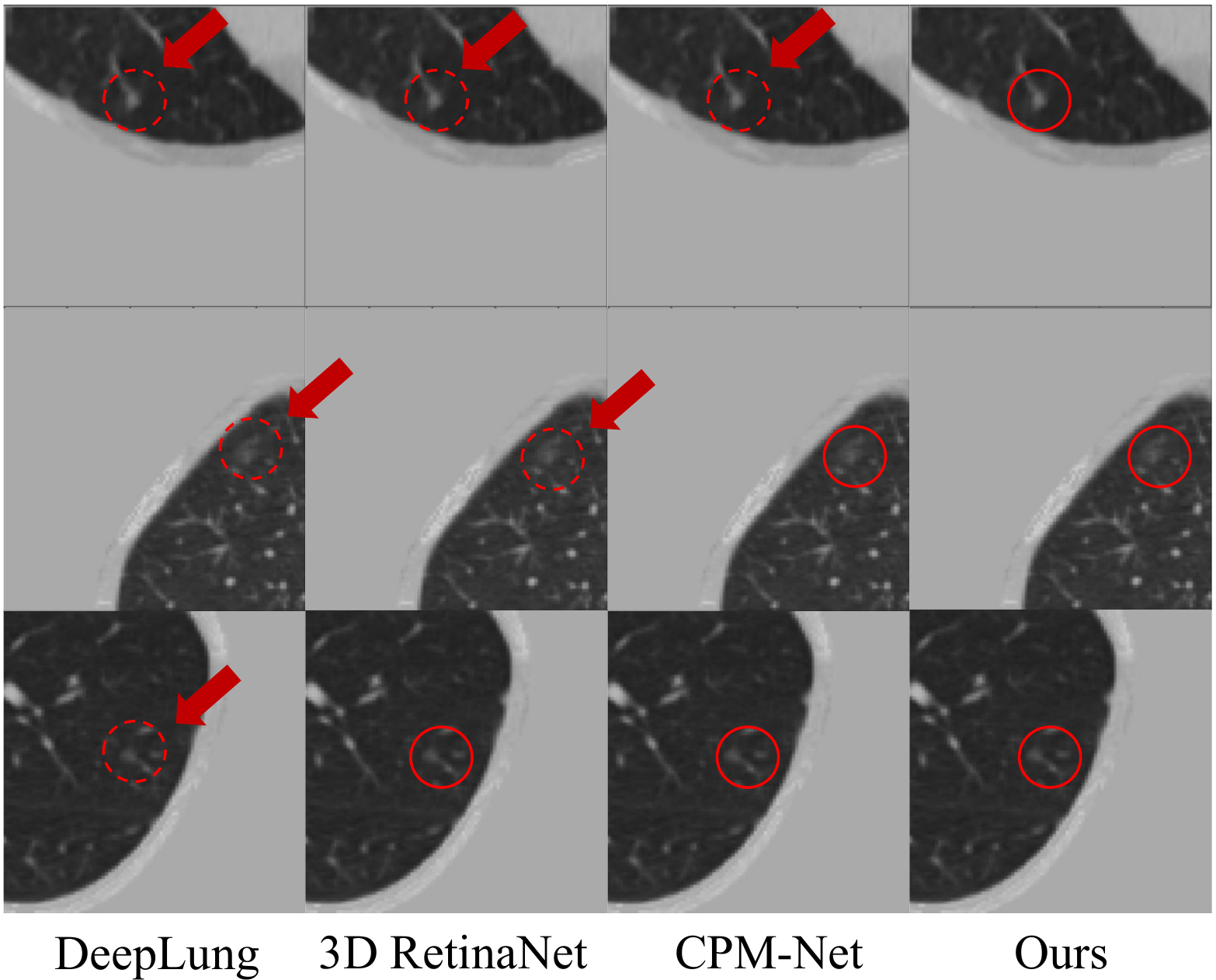}
    \caption{Visualizations of detection results obtained by different methods, where dotted circles represent missed cases and solid circles represent correct detection.}
    \label{fig:visual_2d}
\end{figure}

\begin{figure*}[t]
    \centering
    \includegraphics[width=0.8\textwidth]{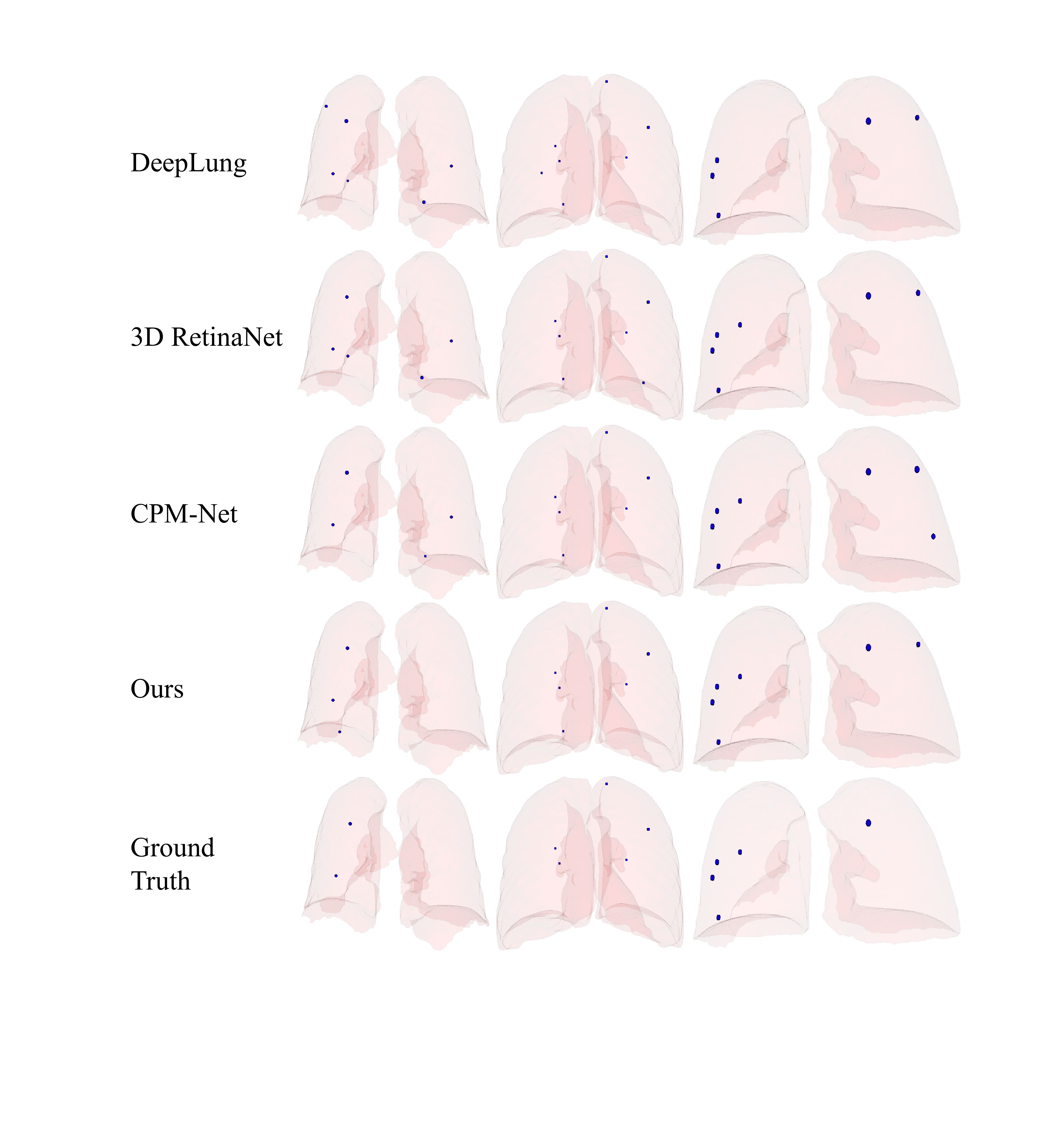}
    \caption{Visualizations of detection results obtained by different methods in the 3D space, where blue spheres represent lung nodules. Due to some lung nodules are too small, we dilate these nodules with a 3 $\times$ 3 $\times$ 3 kernel for better visualization.}
    \label{fig:visual_3d}
\end{figure*}

We compared our method with some powerful anchor-based detectors including Deeplung~\citep{zhu2018deeplung}, 3D RetinaNet~\citep{lin2017focal} and 3D RetinaNet++~\citep{lin2017focal}, and also  with several existing anchor-free-based detectors including 3D CenterNet~\citep{duan2019centernet} and  CPM-Net~\citep{song2020cmpnet}. In this study, we re-implemented the 3D version of CenterNet~\citep{duan2019centernet} and RetinaNet~\citep{lin2017focal} following the original works' descriptions. It should be noted that 3D RetinaNet++ is an enhanced version of 3D RetinaNet, adopting the same OHEM and re-focal loss as our SCPM-Net and using an online anchors' matching strategy instead of offline pre-determining strategy in~\citep{lin2017focal}. We drew the FROC curve in Figure~\ref{fig:froc} demonstrating the sensitivities at seven predefined FPs/Scan rates: $1/8$, $1/4$, $1/2$, $1$, $2$, $4$ and $8$. A quantitative evaluation of these methods at 1,2 and 8 FPs/Scan are listed in Table~\ref{tab:table_cpr}. Our SCPM-Net achieved sensitivities of 92.3\% at 1 FPs/Scan and 92.7\% at 2 FPs/Scan. The anchor-free 3D CenterNet obtained the lowest performance, which is due to that it uses a single centroid to represent a nodule, resulting in location failure and  false negatives. In contrary to CenterNet~\citep{duan2019centernet}, our SCPM-Net uses multi-point matching to reduce the false negatives, and thus increases the sensitivity. Compared with CPM-Net~\citep{song2020cmpnet}, SCPM-Net outperformed CPM-Net in terms of all the three evaluation metrics, since it used $L_{SIoU++}$ for model training which takes many geometric properties into account, i.e., sphere overlap, center point distance and angle of intersection of two spheres. In addition, we also investigated the inference time of each detector with the same experimental settings and hardware. For all one-stage detectors (3D CenterNet, 3D RetinaNet, 3D RetinaNet++, CPM-Net, and SCPM-Net) based on the same backbone, they spend very close inference time (8.37s per scan), but SCPM-Net achieves a better sensitivity than the others. The multi-stage detectors are time-consuming, e.g., DeepLung requires more than 11s per scan for inference. In contrast, SCPM-Net requires less inference time with a higher accuracy.Figure~\ref{fig:visual_2d} and Figure~\ref{fig:visual_3d} show some visualization results of different detectors in the 2D slice-level and 3D volume-level respectively. It can be observed that SCPM-Net is able to capture small nodules that would be missed by the other detectors with fewer false-positive samples.

\begin{table}[tb]
\label{table_cpr}
\centering

\scalebox{0.6}{\begin{tabular}{cccc}
\hline
\multirow{2}{*}{Method}&  \multicolumn{3}{c}{Sensitivity}     \\ \cline{2-4} 
                        & FPs/Scan=1 & FPs/Scan=2 & FPs/Scan=8 \\ \hline
3D CenterNet~\citep{duan2019centernet}            & 71.4\%     & 75.8\%     & 77.5\%     \\ \hline
DeepLung~\citep{zhu2018deeplung}                & 85.8\%     & 86.9\%     & 87.5\%     \\ \hline
3D RetinaNet~\citep{lin2017focal}              & 83.6\%     & 86.8\%     & 88.7\%     \\ \hline
3D RetinaNet++~\citep{lin2017focal}              & 88.8\%     & 91.3\%   & 91.4\%     \\ \hline
CPM-Net~\citep{song2020cmpnet}            & 91.2\%     & 92.4\%     & 92.5\%     \\ \hline

Ours            & \textbf{92.3\%}     & \textbf{92.7\%}     & \textbf{94.8\%}     \\ \hline
\end{tabular}}
\caption{Comparison of different detectors on LUNA16 dataset. (FPs/Scan: the false positives per scan)}
\label{tab:table_cpr}
\end{table}

\subsubsection{Comparison with State-of-the-Art Lung Nodule Detectors}\label{exp:exp5}We further compared our method with several published state-of-the-art lung nodule detection methods on the LUNA16 dataset~\citep{setio2017validation}:~\citet{dou2017automated} employed a two-stage framework which consists of a candidate screening stage and false positive reduction stage;~\citet{zhu2018deeplung} used a three-stage framework consisting of candidate generation, feature extraction and classification;~\citet{liao2019evaluate} presented a two-stage 3D network for lung nodule detection and benign/malignant diagnosis and won the Data Science Bowl 2017 competition;~\citet{tang2019nodulenet} proposed a multi-task framework to solve nodule detection, false-positive reduction, and nodule segmentation jointly;~\citet{song2020cmpnet} developed an anchor-free detector which obtains the detection results by regressing the bounding box size of the nodule and local offset of the center points;~\citet{li2020deepseed} proposed a 3D encoder-decoder network with a ``Squeeze-and-Excitation" attention module and a dynamically scaled cross-entropy loss for lung nodule detection and false-positive reduction at the same time;~\citet{mei2021sanet} proposed a slice-aware network to capture long-range dependencies among any positions and
any channels of one slice group in the feature map. The quantitative comparison of these methods based on a standard 10-fold cross-validation is presented in Table~\ref{tab:sota-comparison}. It shows that our SCPM-Net achieves better performance just with a simpler one-stage framework. Note that we used these works' reported results rather than our re-implemented results in Table~\ref{tab:sota-comparison}, but it would be of interest to compare them with the same software and hardware environment in the future.

% Note that we used these works' reported results rather than we re-implemented results. It shows that the results of our method are better, but it would be of interest to compare them with the same software and hardware environment in the future.} 
% Compared with these state-of-the-art detection frameworks, our SCPM-Net achieves better performance and shows an advantage over the others as it's an efficient end-to-end single-stage detection network.

% SCPMNet:0.125（74.3）0.25（82.9）  0.5（88.9） 1（92.2）2（93.9）4（95.8）8（96.4）avg 89.2
% CPMNet 0.125（72.3）0.25（83.8） 0.5（88.7） 1（91.1）2（92.8）4（93.4）8（94.8）average 88.1
% 0.125 0.25 0.5 1.0 2.0 4.0 8.0
\begin{table*}[t]
\centering
\scalebox{0.65}{\begin{tabular}{ccccccccccc}
\hline
\multirow{2}{*}{Method} & \multicolumn{8}{c}{Sensitivity} & \multirow{2}{*}{FPR} & \multirow{2}{*}{Stages}    \\ \cline{2-9} 
& FPs/Scan=0.125 & FPs/Scan=0.25 & FPs/Scan=0.5 & FPs/Scan=1.0 & FPs/Scan=2.0 & FPs/Scan=4.0 & FPs/Scan=8.0 & Average& & \\ \hline
\citet{dou2017automated}&65.90& 74.50& 81.90& 86.50& 90.60& 93.30& 94.60& 83.90&w&3\\ \hline
\citet{zhu2018deeplung}& 66.20& 74.60& 81.50& 86.40& 90.20& 91.80& 93.20& 83.40&w & 3\\ \hline
\citet{liao2019evaluate} & 59.38& 72.66& 78.13& 84.38& 87.50& 89.06& 89.84& 80.13 &w/o & 2\\ \hline
\citet{tang2019nodulenet}& 70.82& 78.34& 85.68& 90.01& \textbf{94.25}& 95.49& 96.29& 87.27&w&3\\ \hline
\citet{li2020deepseed} & 
73.90& 80.30& 85.80& 88.80& 90.70& 91.60& 92.00& 86.20&w&3\\ \hline

\citet{song2020cmpnet}&72.30 &\textbf{83.80} &88.70 &91.10 &92.80 &93.40 &94.80 &88.10 & w/o & 1 \\ \hline
\citet{mei2021sanet} & 71.17& 80.18& 86.49& 90.09& 93.69& 94.59& 95.50 &87.39& w & 3 \\ \hline 
Ours & \textbf{74.3}&82.9&\textbf{88.9}&\textbf{92.2}&93.9&\textbf{95.8}&\textbf{96.4}&\textbf{89.2}&   w/o  & \textbf{1} \\
\hline
\end{tabular}}
\caption{Comparison between our method and the state-of-the-art methods on LUNA16 dataset using a standard 10-fold cross validation. (FPs/Scan: False positives per scan, FPR: False positive reduction)}
\label{tab:sota-comparison}
\end{table*}

\section{Discussion and Conclusion}
We have successfully evaluated the proposed SCPM-Net and $L_{SIoU++}$ on a large lung nodule detection dataset~\citep{setio2017validation}. To overcome the drawbacks of anchor-based approaches, we propose an anchor-free framework to detect lung nodules in 3D CT scans. To capture context efficiently, we integrate Squeeze-and-Excitation modules~\citep{hu2018squeeze} and multi-level spatial coordinate maps into an encoder-decoder structure for better performance. Inspired by anchor-matching methods~\citep{zhang2019freeanchor, ren2015faster}, we propose a center points matching method to train the detector more efficiently. Based on the fact of annotating nodules as spheres in clinical practice~\citep{macmahon2017guidelines}, we use bounding spheres to represent nodules rather than bounding boxes, and further combine geometric measures to propose a sphere-based intersection-over-union loss function ($L_{SIoU++}$) to train the detector. All these sub-modules and loss functions constitute a powerful and efficient detector for lung nodule detection. In this work, we just used the $L_{SIoU++}$ function to train our proposed anchor-free detectors for lung nodules detection, it may also be extended to train anchor-based detectors.
\par Recently,~\citet{yang2020circlenet} proposed an anchor-free detection method with circle
representation (i.e., CircleNet), and the main differences between
SCPM-Net and CircleNet are:
1) CircleNet uses 2D circles to represent the glomerulus in 2D pathology images, while our method uses spheres to represent  lung nodules in 3D space, which is inspired by clinical practice of lung nodule diagnosis and measurement; 2) For detector optimization, our SCPM-Net uses a sphere-based loss function, and CircleNet just uses circle IoU as a metric for loss calculation. 3) To accelerate the convergence, SCPM-Net introduces many geometric metrics to  $L_{SIoU++}$, which was not considered in CircleNet. Note that our framework can be easily extended to other detection tasks, such as lesion detection, as the SCPM-Net is
a general anchor-free framework and does not rely on any specific task, and the $L_{SIoU++}$ considers many geometric metrics that may be applied to other structures as well. In the future, we will investigate further improving the efficiency of the anchor-free detector, and validate it with other clinical applications. 
\par In conclusion, we propose a novel 3D sphere representation-based center-points matching detection network (SCPM-Net) for pulmonary nodule detection from volumetric CT images. Meanwhile, we use an attentive module consisting of coordinate attention and squeeze-and-excitation attention to capturing spatial position. Besides, we adopt a hybrid method of online hard example mining (OHEM) and re-focal loss to solve the imbalance between positive points and negative points. In addition, we introduce a novel sphere representation for lung nodules detection in 3D space and propose a novel loss function $L_{SIoU++}$ that considers geometric measures for  training. Experimental results on the LUNA16 nodule detection dataset show that the proposed SCPM-Net achieves a better or comparable performance with a simpler architecture compared with several state-of-the-art single-stage and multi-stage approaches. The high sensitivities in single-stage inference demonstrate promising potential for further clinical use.

\section{Acknowledgment}
This work was supported by the National Natural Science Foundations of China [81771921, 61901084], and also by key research and development project of Sichuan province, China [20ZDYF2817]. We would like to thank Mr. Yechong Huang from SenseTime Research for constructive discussions, suggestion and manuscript proofread and also thank the organization teams of LUNA16 challenge for the publicly available datasets.

\bibliographystyle{model2-names.bst}\biboptions{authoryear}
\bibliography{egbib}
\end{document}